\documentclass[letterpaper, 10pt, conference]{ieeeconf}
\IEEEoverridecommandlockouts 
\usepackage{amsmath,amssymb,url}
\usepackage{graphicx,subfigure}
\usepackage{verbatim}
\usepackage{color}
\usepackage{siunitx}
\usepackage[hidelinks]{hyperref}
\usepackage{lipsum} 
\usepackage{cleveref}
\usepackage{booktabs}
\usepackage{algorithm}
\usepackage{algpseudocode}
\usepackage{caption}
\usepackage{adjustbox}
\usepackage{multirow}
\usepackage{makecell} 
\usepackage[table,xcdraw]{xcolor}

\newcolumntype{?}{!{\vrule width 1pt}}

\newcommand{\SO}{\ensuremath{\mathsf{SO(3)}}}

\newcommand{\so}{\ensuremath{\mathfrak{so}(3)}}

\renewcommand{\Re}{\ensuremath{\mathbb{R}}}
\newcommand{\Sph}{\ensuremath{\mathsf{S}}}

\newcommand{\E}{\mathbb{E}}

\title{\LARGE \bf
	Multi-Agent Reinforcement Learning for \\ the Low-Level Control of a Quadrotor UAV
}
\author{Beomyeol Yu and Taeyoung Lee
	\thanks{Beomyeol Yu, Taeyoung Lee, Mechanical and Aerospace Engineering, George Washington University, Washington, DC 20051, {\tt \{yubeomyeol,tylee\}@gwu.edu}}%
    \thanks{\textsuperscript{\footnotesize\ensuremath{*}}This research has been supported in part by NSF CNS-1837382, AFOSR MURI FA9550-23-1-0400, and ONR N00014-23-1-2850.}
}

\graphicspath{{./figs/}}

\makeatletter
\def\endthebibliography{%
	\def\@noitemerr{\@latex@warning{Empty `thebibliography' environment}}%
	\endlist
} 
\makeatother

\begin{document}
	\allowdisplaybreaks
	\maketitle \thispagestyle{empty} \pagestyle{empty}

	\begin{abstract}
		By leveraging the underlying structures of the quadrotor dynamics, we propose multi-agent reinforcement learning frameworks to innovate the low-level control of a quadrotor, where independent agents operate cooperatively to achieve a common goal.
		While single-agent reinforcement learning has been successfully applied in quadrotor controls, training a large monolithic network is often data-intensive and time-consuming.
		Moreover, achieving agile yawing control remains a significant challenge due to the strongly coupled nature of the quadrotor dynamics.
		To address this, we decompose the quadrotor dynamics into translational and yawing components and assign collaborative reinforcement learning agents to each part to facilitate more efficient training.
		Additionally, we introduce regularization terms to mitigate steady-state errors and prevent excessive maneuvers.
		Benchmark studies, including sim-to-sim transfer verification, demonstrate that our proposed training schemes substantially improve the convergence rate of training, while enhancing flight control performance and stability compared to traditional single-agent approaches.
	\end{abstract}

	\section{Introduction}
	Deep reinforcement learning has emerged as a powerful strategy for quadrotor low-level control by enabling direct mapping of high-dimensional sensory observations to control signals.
	This paradigm has opened up new possibilities across various applications, including position tracking~\cite{hwangbo2017control}, drone racing~\cite{song2021autonomous}, and payload transportation~\cite{belkhale2021model}.
	However, the majority of achievements in this field have been centered on single-agent reinforcement learning (SARL) due to its conceptual simplicity and familiarity. 
	
	Training monolithic neural networks for complex control tasks often requires a large amount of training data and computational time. 
	This challenge is particularly pronounced in quadrotor control due to its inherently nonlinear and underactuated nature.
	This is because a single agent is supposed to simultaneously achieve multiple control objectives that exhibit distinct characteristics, relying on the indirect information from the reward function.
	Consequently, prior studies often omitted the heading control, despite its importance in certain applications such as video imaging, as excessive yaw maneuvers may have detrimental effects on position stabilization and tracking performance.
	
	In contrast, multi-agent reinforcement learning (MARL) provides significant potential for achieving complex control goals by breaking them down into smaller subgoals. 
	This collaborative approach fosters faster learning and superior performance through information sharing, further improving system robustness against the failure of individual agents. 
	
	In this work, we propose MARL frameworks for the low-level control of a quadrotor by decomposing its dynamics into the translational part and the yawing part according to the decoupled yaw control strategy proposed in \cite{gamagedara2019geometric}.
	Specifically, in the quadrotor dynamics, the three-dimensional attitude does not have to be controlled completely for position tracking.
	This is because the yawing control, which corresponds to the rotation around the thrust vector, is irrelevant to the translational dynamics. 
	Thus, the position tracking task is supplemented with the control of the yawing direction to determine the complete desired attitude~\cite{lee2010geometric}.
    In this paper, we utilize two agents dedicated to each part of the decomposed dynamics for efficient learning and robust execution.
    
	The primary contributions of this paper are fourfold: 
	(\romannumeral 1) our framework significantly improves the flight stability and robustness of the system by decoupling yaw control from position control, particularly beneficial for large yaw angle flights,
	(\romannumeral 2) the proposed frameworks reduce the overall training time by specializing each agent in its designated task, converging faster with less training data compared to a complex single agent,
	(\romannumeral 3) two regularization terms are introduced to mitigate steady-state errors in position tracking and prevent excessive control inputs, and 
	(\romannumeral 4) our proposed approaches remain flexible and can be trained with any MARL algorithm. 
	Specifically, in this paper, we implement the proposed MARL frameworks with Multi-Agent TD3 (MATD3)~\cite{ackermann2019reducing} and benchmarked it against the single-agent counterpart, Twin Delayed Deep Deterministic Policy Gradient (TD3)~\cite{fujimoto2018addressing}, demonstrating the substantial advantages in the training efficiency.

	\begin{figure*}[t]
		\centering
		\hspace*{0.7cm}
		\begin{picture}(550,180)
			\put(0,0){\includegraphics[width=0.92\textwidth]{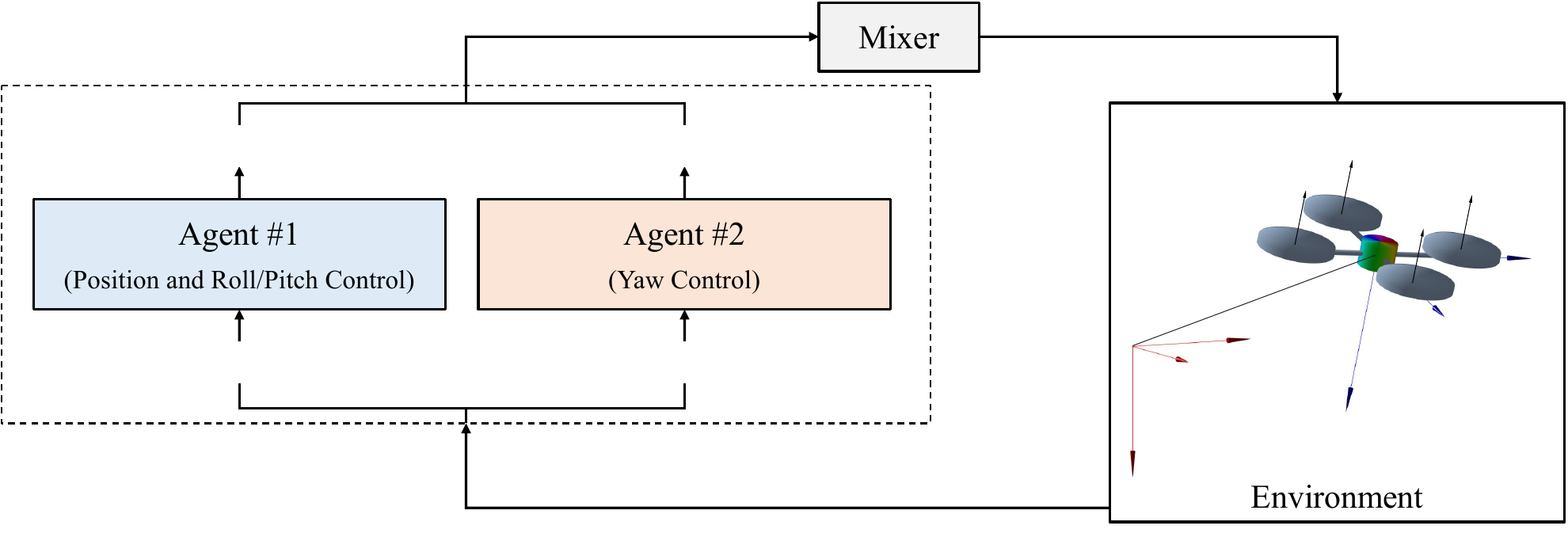}}
			\put(38,113){{\small $a_1 = (f, \tau) \in \Re^4$}}
			\put(177,113){{\small $a_2 = M_3 \in \Re$}}
			
			\put(155,152){{\small $\textrm{Joint action} ~ (a_1, a_2) $}}
			\put(304,152){{\small $(T_1, T_2, T_3, T_4) \in \Re^4$}}
			
			\put(12,50){{\small $o_1=(e_x, e_{I_x}, e_v, b_3, e_{\omega_{12}}) \in \Re^{15}$}}
            \put(168,50){{\small $o_2=(b_1, e_{\Omega_3}) \in \Re^4$}}
			
			\put(150,-3){{\small $\textrm{Global state} ~ s=(x, v, R, \Omega) \in \Re^{9} \times \SO$}}
			\put(142,24){{\small $\textrm{Local observations} ~ (o_1, o_2)$}}
			
			\put(373,56){{\small $\vec e_1$}}
			\put(354,47){{\small $\vec e_2$}}
			\put(334,9){{\small $\vec e_3$}}		
			\put(456,79){{\small $\vec b_1$}}
			\put(428,54){{\small $\vec b_2$}}
			\put(397,27){{\small $\vec b_3$}}
			\put(434,103){{\small $T_1$}}
			\put(418,94){{\small $T_2$}}
			\put(384,105){{\small $T_3$}}
			\put(398,115){{\small $T_4$}}
		\end{picture}
		\vspace*{0.01mm}
		\caption{MARL structure for the low-level control of a quadrotor:
			Agent $\#$1 controls the roll/pitch dynamics and the position, and Agent $\#$2 controls the yaw.
			Each agent receives individual observations $(o_1, o_2)$ and rewards $(r_1, r_2)$.
			They select the optimal joint action $(a^*_1, a^*_2)$ with their policy networks, which is converted into motor thrust $T_{1:4}$ by the mixer in \eqref{eqn:mixer}.}
		\label{fig:MARL}
		\vspace*{-0.2cm}
	\end{figure*}
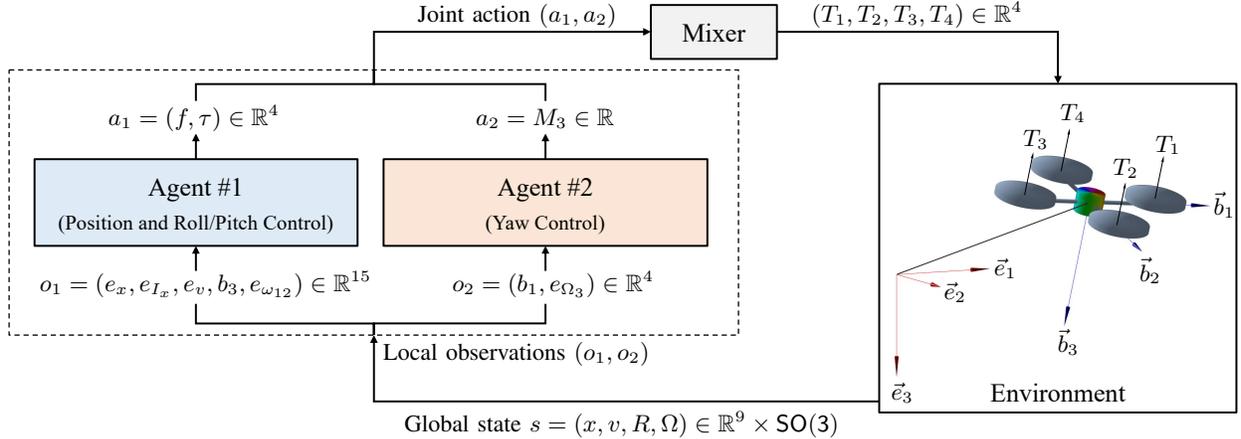

	\section{Related work}\label{sec:RW} 
	
	\textbf{SARL for Quadrotor Control.} 
	Prior works have mainly focused on developing quadrotor control strategies for stabilization and trajectory tracking. 
	The pioneering work in \cite{hwangbo2017control} introduced deep reinforcement learning for quadrotor low-level control.
    Expanding the scope, Proximal Policy Optimization (PPO)-based algorithms were applied for quadrotor position tracking control in \cite{pi2020low}, which was further extended and generalized to multi-rotor aerial vehicles in~\cite{pi2021general}.
	The study presented in \cite{molchanov2019sim} explored sim-to-(multi)-real transfer of low-level control policies, bridging the \textit{reality gap}. 
	They successfully trained and transferred a single robust control policy from simulation to multiple real quadrotors. 
    Recently, a data-efficient learning strategy was proposed in \cite{yu2023equivariant} in the context of equivariant reinforcement learning, leveraging the symmetry of the quadrotor dynamics. 

	
	\textbf{MARL for Quadrotor Control.} 
	While multi-agent RL has received significant attention in quadrotor applications, the primary focus has been on higher-level decision-making for multiple quadrotors, including cooperative trajectory planning and formation control while ensuring collision avoidance~\cite{li2021distributed,batra2022decentralized,ourari2022nearest}.
	Notably, a cascade MARL low-level control framework for a quadrotor has been proposed in \cite{han2022cascade} by decomposing quadrotor dynamics into six one-dimensional subsystems.
	However, their approach aims to train multiple single agents independently through traditional SARL algorithms, implying that each subsystem is controlled by a separate agent rather than operating in a cooperative manner. 
    Moreover, they decomposed the dynamics under the assumption of small-angle maneuvers, which limits its applicability to complex scenarios involving large rotations.
	\textbf{Quadrotor Decoupled Yaw Control.} 
	A geometric control system with decoupled yaw was proposed in \cite{gamagedara2019geometric}, where the quadrotor attitude dynamics is decomposed into two distinct components: the roll/pitch part, which controls thrust direction, and the yaw part, responsible for heading direction.
	This was further extended in \cite{gamagedara2022geometric} by integrating adaptive control terms to address uncertainties and disturbances.
	Relying on the Lyapunov stability, these works should be distinguished from the presented data-driven reinforcement learning. 

	In comparison to prior studies, our work presents a unique perspective of utilizing MARL for single quadrotor control by decoupling yaw control from translational control, thereby improving training efficiency and stability as well as control performance against SARL. 


	\section{Backgrounds}\label{sec:BG} 

	\subsection{Single‑Agent Reinforcement Learning} 
	Reinforcement learning is a subset of machine learning in which the decision-maker, called an \textit{agent}, learns to make optimal decisions (or actions) through a process of trial and error. 
	This learning process involves the agent interacting with an environment and occasionally receiving feedback in the form of rewards. 
	The Markov Decision Process (MDP) offers a formal framework for modeling decision-making problems, formalized by a tuple $\mathcal{M} = (\mathcal{S}, \mathcal{A}, \mathcal{R}, \mathcal{T}, \gamma)$, where $\mathcal{S}$ and $\mathcal{A}$ are the state and action spaces, respectively.
	The reward function is denoted by $\mathcal{R}: \mathcal{S} \times \mathcal{A} \rightarrow \mathbb{R}$, and the transition probability function as $\mathcal{T: \mathcal{S} \times \mathcal{A} \rightarrow \mathcal{P}(\mathcal{S})}$ specified by ${s}_{t+1} \sim P({s}_{t+1}|{s}_t, {a}_t)$ in a discrete-time setting.
		
	The behavior of the agent is determined by its policy.
	At each time step $t$, the agent selects its action ${a}_t$ according to a \textit{policy} $\pi({a}_t|{s}_t)$, which is the probability distribution of the action conditioned on the state $s_t$.
	The common objective is to maximize the expected return,
	\begin{align}
		J(\pi) = \E_{{\tau} \sim \pi} [\sum_{t=0}^{T-1} \gamma^t r({s}_t, {a}_t)], \label{eqn:J}	
	\end{align}
	where $\gamma \in [0, 1]$ is a discount factor, and a trajectory ${\tau} = (s_0, a_0, \cdots, s_T,  a_T)$ is a sequence of state-action pairs.
	Thus, the optimal policy $\pi^*$ can be expressed as 
	\begin{align}
		{\pi}^* = \underset{{\pi}}{\arg\max} ~ J(\pi). 
	\end{align}

	Alternatively, the optimality is often formulated through a state-value function $V^\pi : \mathcal{S} \rightarrow \Re$ according to the policy $\pi$,
	\begin{align}
	V^\pi (s) = \E_{{\tau} \sim \pi} [\sum_{t=0}^{T-1} \gamma^t r({s}_t, {a}_t) | s_0 = s] \label{eqn:V_SARL}, 
	\end{align}
	Instead, it is often useful to introduce an action-value function $Q^\pi: \mathcal{S} \times \mathcal{A} \rightarrow \Re$, also known as a Q-value, which corresponds to the value conditioned not only on the current state but also on the chosen action, i.e.,
	\begin{align}
	Q^\pi (s,a) = \E_{{\tau} \sim \pi}  [\sum_{t=0}^{T-1} \gamma^t r({s}_t, {a}_t) | s_0 = s, a_0 = a].
	\end{align}
	In general, $V^\pi (s)$ and $Q^\pi (s, a)$ are constructed and updated by the Bellman equation in a recursive way.
	In the context of deep reinforcement learning, the policy and value functions are designed as neural networks.
	
	\subsection{Multi‑Agent Reinforcement Learning}

	In multi-agent decision-making problems, the partially observable Markov games (POMG), also known as the partially observable stochastic game (POSG), serve as a framework for modeling the interactions between diverse agents within a shared environment. 
	Multiple decision-makers select and execute their actions simultaneously, and each agent's reward and next state are determined by the joint action of all agents. 
	Specifically, the POMG is defined by the tuple $(\mathcal{N}, \mathcal{S}, \{\mathcal{A}_i\}, \{\mathcal{O}_i\}, \{\mathcal{R}_i\}, \{b_i^0\}, \mathcal{T}, \gamma )$ for $i \in \mathcal{N}$, where $\mathcal{N} = (1, \cdots, N)$ is the set of agents and $\mathcal{S}$ is the set of states. 
	Next, $\mathcal{A} = \mathcal{A}_1 \times \cdots \times \mathcal{A}_N$ is the joint action space where $\mathcal{A}_i$ is the action space for $i$-th agent.
	Similarly, $\mathcal{O} = \mathcal{O}_1 \times \cdots \times \mathcal{O}_N$ is the joint observation space where $\mathcal{O}_i$ is a set of observations of agent $i$.
	Also, $\mathcal{R}_i : \mathcal{S} \times \mathcal{A} \rightarrow \Re$ is the reward function of $i$-th agent and $b_i^0$ denotes the initial state distribution of $i$-th agent.
	Here, each agent $i$ chooses its action $a_i$ depending on its individual policy $\pi_i : \mathcal{O}_i \rightarrow  \mathcal{P}(\mathcal{A}_i)$, which yields the next state according to the state transition function $\mathcal{T}$, and receives the local observation correlated with $o_i : \mathcal{S} \rightarrow \mathcal{O}_i$.
	
	A common way to solve POMGs is \textit{Multi-Agent Actor Critic}, proposed in \cite{lowe2017multi}, adopting the Centralized Training with Decentralized Execution (CTDE) framework. 
	Consider $N$ agents, whose policies $\pi = (\pi_1, \cdots, \pi_N)$ are parameterized by ${\theta} = (\theta_1, \cdots, \theta_N)$.
	In this approach, each agent has its own actor network determining a deterministic action $a_i$ from its local observation $o_i$. 
	The joint state $s = (o_1, \cdots, o_N )$ and joint action $a = (a_1, \cdots, a_N )$ correspond to the observations and actions of all agents, respectively.
	The centralized critic network estimates the joint Q-value $Q^\pi_i (s,a)$ based on the joint state-action pairs $(s, a)$, providing Q-value approximations to decentralized actor networks. 
	To train agents, an experience replay buffer $\mathcal{D}$ stores tuples $\tau = (s, s', a, r_1, \cdots, r_N)$ recording transitions for all agents. 
	Then, a batch of these transitions is randomly sampled from the buffer during training.
	Next, the one-step lookahead TD error is computed to update the centralized critic network as
	\begin{align} \label{eqn:MADDPG_critic}
		\mathcal{L}(\theta_i)= \mathbb{E}_{\tau \sim \mathcal{D}} \bigr[ (Q_i^{\pi} (s,a) - y)^2 \bigr],		
	\end{align}	
	where $y = r_i + \gamma Q_i^{\pi'} (s', a') \rvert_{a'_j = {\pi^\prime_j} (o_j) }$ for the set of target policies $\pi'$. 
	Lastly, the deterministic policy gradient updates each agent’s actor parameters as
	\begin{align} \label{eqn:MADDPG_actor}	
		\nabla_{\theta_i} J(\pi_i) = \mathop{\mathbb{E}}_{\tau \sim \mathcal{D}} \bigr[ \nabla_{\theta_i}\pi_i (a_i|o_i) \nabla_{a_i} Q^{\pi}_i (s, a) \rvert_{a_i = {\pi_i} (o_i)} \bigr].
	\end{align}	

	Importantly, if the actions of all agents are known, the environment remains stationary even as policies change, addressing \textit{non-stationarity}, since $P(s'|s,a,\pi_1,\cdots,\pi_N) = P(s'|s,a) = P(s'|s,a,\pi'_1,\cdots,\pi'_N)$ for any $\pi_i \neq \pi'_i$. 
	
	\subsection{Quadrotor Dynamics} 
	Consider a quadrotor UAV, illustrated at \Cref{fig:MARL}, where $\{\vec e_1, \vec e_2, \vec e_3\}$ corresponds to the inertial frame, and
	$\{\vec b_{1}, \vec b_{2}, \vec b_{3}\}$ denotes the body-fixed frame located at the mass center of the quadrotor.
	The position, velocity, and angular velocity of the quadrotor are represented by $x \in \Re^3$, $v \in \Re^3$, and $\Omega = [\Omega_1, \Omega_2, \Omega_3]^T \in \Re^3$, respectively, where $x$ and $v$ are resolved in the inertial frame, and $\Omega$ is in the body-fixed frame. 
    The attitude is defined by the rotation matrix $R = [b_1, b_2, b_3] \in \SO=\{R\in\Re^{3\times 3}\,|\, R^T R=I_{3\times 3},\; \mathrm{det}[R]=1\}$, where $b_i\in\Re^3$ is the representation of $\vec b_i$ resolved in the inertial frame for $i\in\{1,2,3\}$. 

    From \cite{lee2010geometric}, the equations of motion are given by 
	\begin{gather}
		\dot x  = v,\label{eqn:x_dot}\\
		m \dot v = mge_3 - f R e_3,\label{eqn:v_dot}\\
		\dot R = R\hat\Omega,\label{eqn:R_dot}\\
		J\dot \Omega + \Omega\times J\Omega = M,\label{eqn:W_dot}
	\end{gather}
	where the \textit{hat map} $\hat\cdot:\Re^3\rightarrow\so=\{S\in\Re^{3\times 3}\,|\, S^T = -S\}$ is defined by the condition that $\hat x y=x\times y$, and $\hat x$ is skew-symmetric for any $x,y\in\Re^3$.
	The inverse of the hat map is denoted by the \textit{vee map} $\vee:\so\rightarrow\Re^3$.
	Next, $m\in\Re$ and $J\in\Re^{3\times 3}$ are the mass and the inertia matrix of the quadrotor with respect to the body-fixed frame, respectively, and  $g\in\Re$ is the gravitational acceleration.
	From the total thrust $f = \sum_{i=1}^{4} T_i \in\Re$ and the moment $M = [M_1, M_2, M_3]^T \in\Re^3$ resolved in the body-fixed frame, the thrust of each motor can be computed by
	\begin{gather} \label{eqn:mixer}
		\begin{bmatrix} 
			T_1 \\ T_2 \\ T_3 \\ T_4
		\end{bmatrix}
		= \frac{1}{4}
		\begin{bmatrix}
			1 & 0      & 2/d   & -1/c_{\tau f} \\
			1 & -2/d & 0      & 1/c_{\tau f} \\
			1 & 0      & -2/d & -1/c_{\tau f} \\
			1 & 2/d   & 0      & 1/c_{\tau f} 
		\end{bmatrix}
		\begin{bmatrix} 
			f \\ M_1 \\ M_2 \\ M_3 
		\end{bmatrix}.
	\end{gather}
	where $d\in\Re$ is the distance between the center of the rotor and the third body-fixed axis, and $c_{\tau f} \in\Re$ is a coefficient relating the thrust and the reactive torque. 
	
	\begin{figure*}[t]
		\centering
		\hspace*{0.7cm}
		\begin{picture}(550,110)
		\put(0,0){\includegraphics[width=0.92\textwidth]{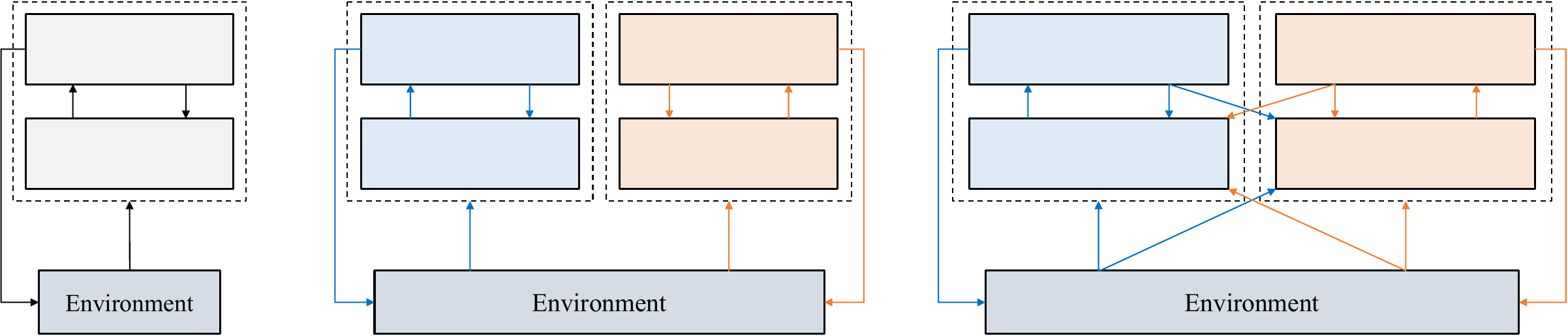}}
		\put(19,82){{\small $\pi^\textrm{SARL} (s ; \theta)$}}
		\put(13,51){{\small $Q^\textrm{SARL} (s, a ; \phi)$}}
		\put(-20,45){{\footnotesize $a^\textrm{SARL}$}}
		\put(18,26){{\footnotesize $s^\textrm{SARL}~r^\textrm{SARL}$}}
		
		\put(116,82){{\small $\pi_1^\textrm{DTDE} (o_1 ; \theta_1)$}}
		\put(193,82){{\small $\pi_2^\textrm{DTDE} (o_2 ; \theta_2)$}}	
		\put(108.5,51){{\footnotesize $Q_1^\textrm{DTDE} (o_1, a_1 ; \phi_1)$}}
		\put(185.2,51){{\footnotesize $Q_2^\textrm{DTDE} (o_2, a_2 ; \phi_2)$}}
		\put(78,45){{\footnotesize $a_1^\textrm{DTDE}$}}
		\put(247.5,1){{\footnotesize $a_2^\textrm{DTDE}$}}
		\put(118,26){{\footnotesize $o_1^\textrm{DTDE}~r_1^\textrm{DTDE}$}}
		\put(195,26){{\footnotesize $o_2^\textrm{DTDE}~r_2^\textrm{DTDE}$}}
		
		\put(302.5,82){{\small $\pi_1^\textrm{CTDE} (o_1 ; \theta_1)$}}
		\put(393,82){{\small $\pi_2^\textrm{CTDE} (o_2 ; \theta_2)$}}	
		\put(290,51){{\footnotesize $Q_1^\textrm{CTDE} (o_{1:2}, a_{1:2} ; \phi_1)$}}
		\put(381,51){{\footnotesize $Q_2^\textrm{CTDE} (o_{1:2}, a_{1:2} ; \phi_2)$}}
		\put(280,21){{\footnotesize $a_1^\textrm{CTDE}$}}
		\put(444,24){{\footnotesize $a_2^\textrm{CTDE}$}}
		\put(304.5,30){{\footnotesize $o_1^\textrm{CTDE}~r_1^\textrm{CTDE}$}}
		\put(396.5,30){{\footnotesize $o_2^\textrm{CTDE}~r_2^\textrm{CTDE}$}}
		\end{picture}
		
		\caption{Three different training schemes in single and multi-agent settings for quadrotor control tasks.
			(left) Single-Agent Reinforcement Learning (SARL) holds a large end-to-end policy that directly outputs motor thrust.
			(middle) In Decentralized Training with Decentralized Execution (DTDE) setting, each agent updates its own policy without explicit information exchange with each other.
			(right) Centralized Training with Decentralized Execution (CTDE) allows agents to share experience during training, but each action is executed decentrally based on their local observations and policies.}
		\label{fig:FW}
		\vspace*{-0.2cm}
	\end{figure*}
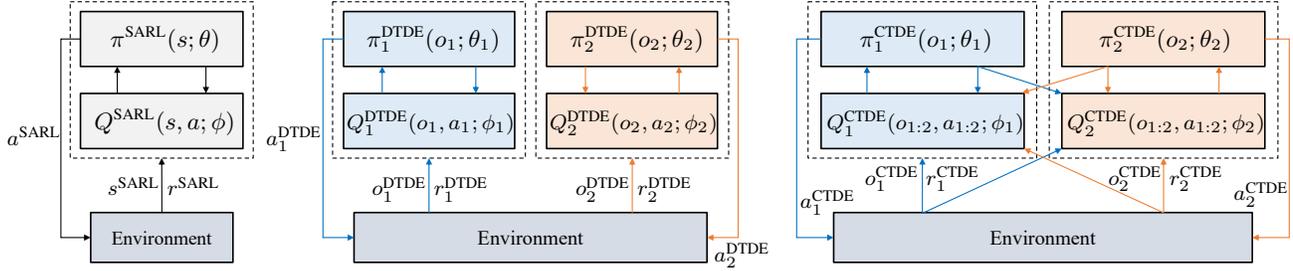

	\subsection{Decoupled Yaw Control System}  \label{sec:BG_decoupled}
	In \eqref{eqn:x_dot} and \eqref{eqn:v_dot}, the translational dynamics are coupled with the attitude dynamics only through $Re_3 = b_3$.
    In other words, the direction of $b_1$ and $b_2$ does not affect the translational dynamics. 
    This is because the direction of thrust is fixed to $b_3$ always, and any rotation about $b_3$ does not affect the resultant thrust. 
    In fact, the dynamics of $b_3$ can be separated from the full attitude dynamics \eqref{eqn:R_dot} and \eqref{eqn:W_dot}, by assuming that the quadrotor is inertially symmetric about its third body-fixed axis, i.e., $J_1 = J_2$.
	This yields the following translational dynamics decoupled from the yaw,
	\begin{gather}
		\dot x  = v,\label{eqn:decoup_x_dot}\\
		m \dot v = mge_3 - f b_3,\label{eqn:decoup_v_dot}\\
		\dot b_3 = \omega_{12} \times b_3,\label{eqn:decoup_b3_dot}\\
		J_1 \dot \omega_{12} = \tau,\label{eqn:decoup_w12_dot}
	\end{gather}
	See \cite{gamagedara2019geometric, gamagedara2022geometric} for details.
	Here, $\omega_{12}\in\Re^3$ is the angular velocity of $b_3$ resolved in the inertial frame, i.e.,
		$\omega_{12} = \Omega_1 b_1 + \Omega_2 b_2$, 
    satisfying $\omega_{12}\cdot b_3 = \dot\omega_{12}\cdot b_3 = 0$.
    In \eqref{eqn:decoup_w12_dot}, the fictitious control moment is defined by $\tau = \tau_1 b_1 + \tau_2 b_2 \in \Re^3$ where $\tau_1 =  M_1 -J_3\Omega_2\Omega_3\in\Re$ and $\tau_2 = M_2 + J_3\Omega_3\Omega_1\in\Re$, which are related to the first two components of the actual moment $(M_1,M_2)$.
	While there are three elements in $\tau$, it has two degrees of freedom, as $\tau \cdot b_3=0$ always. 
	
	Next, the remaining one-dimensional yaw dynamics can be written as
	\begin{align}
		\dot b_1 &= \omega_{23}\times b_1 = - \Omega_2 b_3 + \Omega_3 b_2,\label{eqn:decoup_b1_dot}\\
		J_3 \dot \Omega_3 &= M_3,\label{eqn:decoup_W3_dot}
	\end{align}
	where $\omega_{23} = \Omega_2 b_2 + \Omega_3 b_3\in\Re^3$.
    This can be further reduced into a double-integrator by formulating a yaw angle. 
	Let a reference direction of the first body-fixed axis be a smooth function $b_{1_d} \in \Sph^2$. 
	It is projected into the plane normal to $b_3$ to obtain a projected yaw command $b_{1_c} = (I_{3\times 3}- b_3 b_3^T ) b_{1_d}$.
    Let the yaw angle $e_{b_1}\in [0,\pi)$ be the angle between $b_1$ and $b_{1_c}$.
	Also, let $\omega_c\in\Re^3$ be the angular velocity of $b_{1_c}$ when resolved in the inertial frame, i.e., $\dot b_{1_c} = \omega_c\times b_{1_c}$. 
	Then yawing dynamics can be represented by
	\begin{align}
        J_3 \ddot{e}_{b_1} & = M_3 - J_3 \dot{\omega}_{c_3}. \label{eqn:decoup_psi_2dot}
	\end{align}

	In short, the quadrotor dynamics is decomposed into the translational dynamics controlled by three-dimensional control inputs $(f, \tau_1, \tau_2)$, and the yaw dynamics controlled by the third moment $M_3$.

	\section{Multi-Agent RL for Quadrotor Control} \label{sec:MA_quad} 
    Given a goal state $g$ that consists of the desired position $x_d\in\Re^3$, desired velocity $v_d\in\Re^3$, desired heading direction $b_{1_d}\in\Sph^2$, and desired angular velocity $\Omega_d\in\Re^3$, we aim to find an optimal policy such that $s_t \rightarrow g_t$  as $t\rightarrow\infty$ from arbitrary initial states.
    During training, the goal state is assumed to be fixed, i.e., $x_d=v_d = \Omega_d =[0,0,0]^T$ along with $b_{1_d}=[1,0,0]^T$. 
	Leveraging error states, we achieve trajectory tracking by shifting the goal state.

    \subsection{Single-Agent RL Framework} 
    A single-agent framework directly maps the observation, $o = (e_x, e_{I_x}, e_v, R, e_\Omega) \in \Re^{12}\times \SO$, to the action, $a = (T_{1}, T_{2}, T_{3}, T_{4}) \in \Re^4$.
    Here, $e_x=x-x_d\in\Re^3$, $e_v=v-v_d\in\Re^3$, and $e_\Omega=\Omega-\Omega_d\in\Re^3$ represent the tracking errors of position, velocity, and angular velocity, respectively, while $e_{I_x}\in\Re^3$ is the integral error of $e_x$, to be introduced later in \eqref{eqn:SS}.
	To balance conflicting control objectives, the reward function is defined as 
	\begin{align} 
		r & = -k_x \|e_x\|^2 -k_{I_x} \|e_{I_x}\|^2 -k_{b_1} \|e_{b_1}\| -k_{I_{b_1}} \|e_{I_{b_1}}\|  \nonumber\\
		& \quad -k_{b_3} \|e_{b_3}\| -k_v \|e_v\|^2-k_\Omega \|e_\Omega\|^2 - r_\textrm{crash}.
	\end{align}
	where $k \in\Re$ are positive weighting constants.
	The first four terms aim to minimize position and heading errors, while the subsequent three terms penalize aggressive motions.
	Additionally, a severe penalty $r_\textrm{crash}$ is imposed in the event of a crash or exceeding the predefined flight boundaries.

	\subsection{Multi-Agent RL Frameworks}
    As presented in \Cref{sec:BG_decoupled}, the quadrotor dynamics can be decomposed into translational and yawing parts, each exhibiting distinct dynamic characteristics:
    (\romannumeral 1) The objective in controlling the translational dynamics is to adjust the resultant thrust by changing the thrust magnitude $f$, while rotating the third body-fixed axis $b_3$ to control the direction of the thrust. 
    This is achieved by the difference in thrusts between rotors on opposite sides.
    (\romannumeral 2) The yawing control aims to adjust the heading direction that is irrelevant to the translational dynamics, using the small reactive torque of each rotor. 
    The yawing direction should be carefully adjusted as any aggressive yawing maneuver may amplify the undesired effects caused by large modeling errors in the reactive torque model, potentially jeopardizing the translational dynamics. 
    As such, the translational motion control, which is critical for the overall performance and safety, should be prioritized over the yawing control \cite{gamagedara2019geometric,gamagedara2022geometric}. 

    In the SARL framework, it is challenging to take account of these considerations in the training and execution, as it is supposed to complete both tasks with distinct characteristics simultaneously. 
    To address this, this work proposes assigning a dedicated agent to each controllable part, as depicted in \Cref{fig:MARL}.
	Specifically, the first agent, based on equations \eqref{eqn:decoup_x_dot}--\eqref{eqn:decoup_w12_dot}, is responsible for controlling the translational motion, primarily minimizing position errors. 
	The second agent, based on equations \eqref{eqn:decoup_b1_dot}--\eqref{eqn:decoup_W3_dot}, controls the yaw motion by minimizing heading errors.
	During training, each agent receives local observations $(o_1, o_2)$ and individual rewards $(r_1, r_2)$ based on the control performance of the quadrotor. 
	The agents then independently select their actions $(a_1, a_2)$ according to their policy networks. 
	The resulting joint action $(f, \tau, M_3) \in \Re^5$ is converted to the total force and moments, $(f, M_1, M_2, M_3) \in \Re^4$ where $M_1 =  b_1^T \tau + J_3 W_2 W_3$ and $M_2 =  b_2^T \tau - J_3 W_3 W_1$.
	Lastly, actual motor thrust commands $(T_1, T_2, T_3, T_4) \in \Re^4$ are computed by the mixer in \eqref{eqn:mixer}.
	
	This work presents two multi-agent frameworks depending on the degree of information sharing between the two agents:\\
	\textbf{DTDE Structure.}
	The first scheme involves \textit{Decentralized Training with Decentralized Execution}, in which each agent learns its own policy without any communication with the other agent.
    Each agent perceives its observations, $o_1=(e_x, e_{I_x}, e_v, b_3, e_{\omega_{12}}) \in \Re^{15}$ and $o_2=(b_1, e_{\Omega_3}) \in \Re^4$, respectively, and independently chooses its own action, $a_1 = (f, \tau) \in \Re^4$ and $a_2 = M_3 \in \Re$, to maximize the individual reward  defined by
	\begin{align*}
		r_1 & = -k_x \|e_x\|^2 -k_{I_x} \|e_{I_x}\|^2 -k_v \|e_v\|^2 -k_{b_3} \|e_{b_3}\| \\ 
		~~~~~ & \quad  -k_{\omega_{12}} \|e_{\omega_{12}} \|^2 - r_\textrm{crash}, \\
		r_2 & = -k_{b_1} \|e_{b_1}\| -k_{I_{b_1}} \|e_{I_{b_1}}\| -k_{\Omega_3} \|e_{\Omega_3}\|^2 - r_\textrm{crash}. 
	\end{align*}
	where $e_{b_3} \in (0, \pi]$ is the error between $b_{3_d}$ and third body-fixed axis $b_{3}$ for stabilization. \\
	\textbf{CTDE Structure.}
	In the second strategy, the two agents exchange experience during training but act independently during execution. 
	They cooperate with each other to achieve the desired quadrotor motion by jointly optimizing their actions through \textit{centralized critic networks}. 
	Each agent's centralized critic network takes both agents' observations and actions, $(o_1, a_1, o_2, a_2)$, as input, and outputs each action-value function representing the expected return of the two agents.
	Each agent then updates its policy using its action-value function and makes independent decisions. 
	Note that CTDE and DTDE share the same observations $(o_1, o_2)$, actions $(a_1, a_2)$, rewards $(r_1, r_2)$, and network architectures.
	
    \subsection{Policy Regularization}
	Despite the recent success of RL in quadrotor control, a key challenge remains: trained agents often generate control signals with high-frequency oscillations and discontinuities that are impossible to implement in hardware experiments. 
	This may lead to poor control performance and even hardware failure.
	A common way to address this is reward engineering, where undesirable behavior is mitigated by tuning the parameters of the reward function.
	However, this process is often non-intuitive and does not provide any guarantee that the policy will be trained in the desired way, as the agent learns through indirect information from the reward function. 
	
	
	To address this, \cite{mysore2021regularizing} introduced two regularization terms directly optimizing action policies.
	First, a temporal smoothness term, $L_T = D \bigl( \pi(s_t), \pi(s_{t+1}) \bigl)$, penalizes the policy when subsequent actions differ from previous actions to preserve control smoothness over time $t$.
	Here, $D$ denotes a Euclidean distance measure, i.e., $D(a_1, a_2) = ||a_1 - a_2||_2$.
	Next, a spatial smoothness term, $L_S = D \bigl( \pi(s_t), \pi(\bar{s}_t) \bigl)$, encourages that the selected action does not change drastically when the state is perturbed,
	where the perturbed state $\bar s_t$ is sampled from the normal distribution $\mathcal{N}(s_t, \sigma)$ for $\sigma>0$.
	
	However, we observed that the agents trained with these regularization terms occasionally generate excessive motor thrust in their attempts to rapidly minimize position and yawing errors.
	This is because $L_T$ and $L_S$ do not take into account the magnitude of control signals.
	Thus, we propose a magnitude regulation term to avoid aggressive controls,
	\begin{align}
		L_M = D \bigl(  \pi(s_t),  \bar{a} \bigl), \label{eqn:L_M}
	\end{align}
	where $\bar{a}$ is a nominal action.
	For example, the nominal action of SARL is selected as $\bar{T} = 0.25 mg$, the thrust of each rotor required for hovering.
	In contrast, for DTDE and CTDE, the torques are set to zero, i.e., $\bar{\tau} = \bar{M}_3 = 0$, with $\bar{f} = mg$.
	This not only reduces power consumption but also avoids rapid heading control.
	In addition, by constraining $\pi(s_t)$ around $\bar{a}$, the exploration space of possible actions is restricted to a smaller range.
	This leads to faster learning, as the action policy can focus on a smaller set of possibilities.
	
    Thus, the policy optimization function in \eqref{eqn:J} is augmented with these regularization terms into
	\begin{align}
		\hat{J} (\pi) = J (\pi) -\lambda_T L_T -\lambda_S L_S -\lambda_M L_M. \label{eqn:J_caps}
	\end{align}
	where $\lambda >0 \in\Re$ are regularization weights.
	
    \subsection{Integral Terms for Steady-State Error}
    RL-based quadrotor low-level control often suffers from steady-state errors~\cite{lopes2018intelligent}. 
	To address this and also to improve robustness, an integral term is formulated as
	\begin{align}
        \dot e_{I_x} = -\alpha e_{I_x} + e_x. \label{eqn:SS}
	\end{align}
	with $\alpha >0$ chosen to mitigate the integral windup. 

	\section{Numerical Experiments}\label{sec:EX}
	
	\subsection{Implementation}	
	\begin{figure}[b]
		\centering
		\includegraphics[scale=0.185]{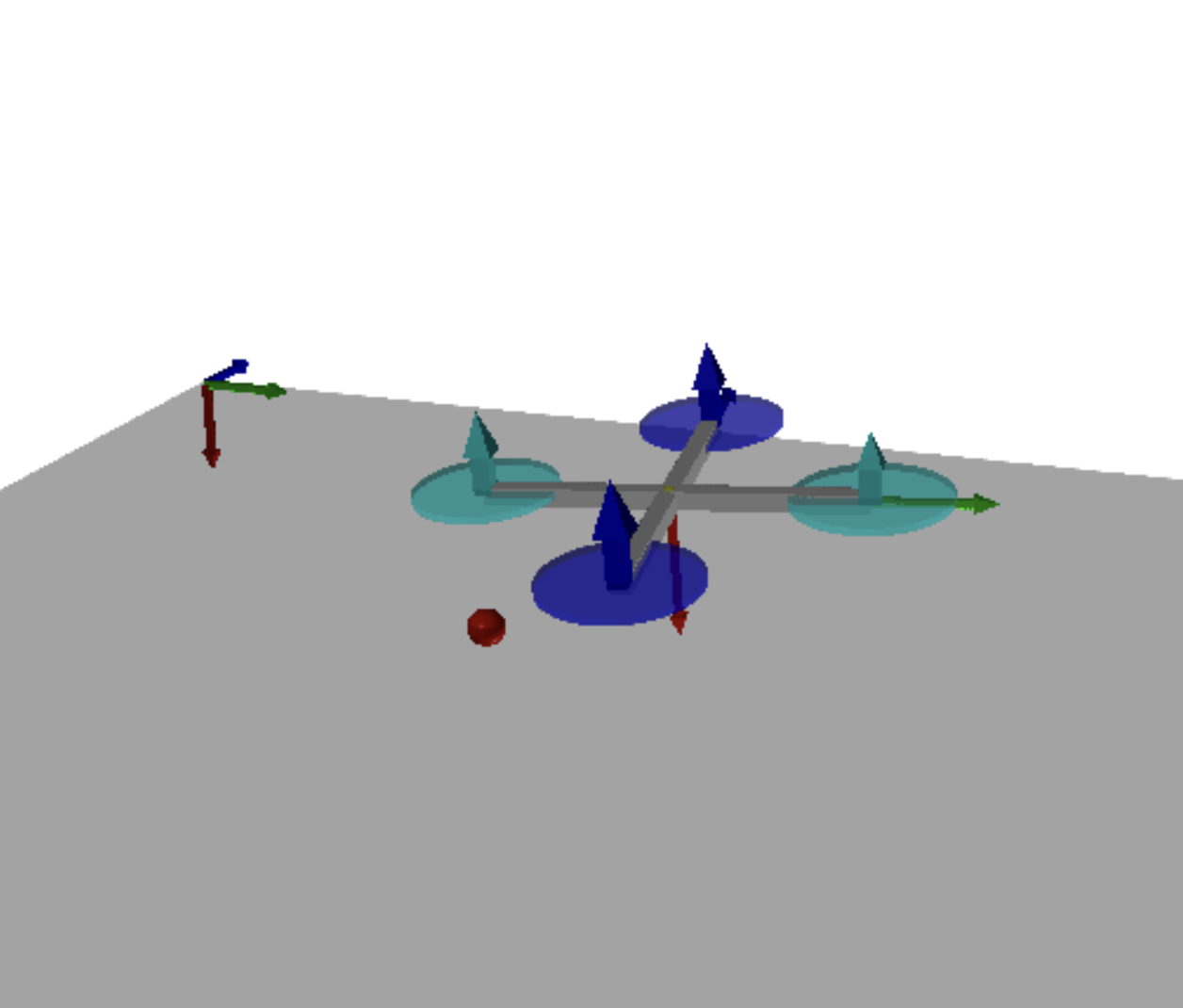} 
		\hfill
		\includegraphics[scale=0.155]{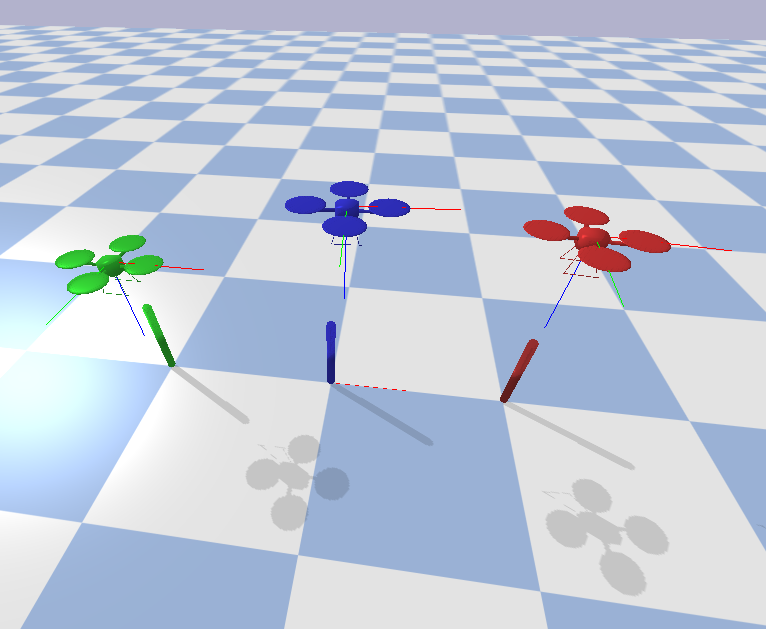}
		\caption{Simulation environments: 
			(left) simplified custom environment for training.
			(right) real-time physics environment, \textit{gym-pybullet-drones}~\cite{panerati2021learning}, for sim-to-sim verification for each scheme of SARL (green), DTDE (blue), and CTDE (red).}
		\label{fig:envs}
	\end{figure}	
	
    We developed a custom training environment based on the equations of motion \eqref{eqn:x_dot}--\eqref{eqn:W_dot} using \textit{OpenAI Gym} library, as shown in \Cref{fig:envs}, where the equations of motion are numerically integrated. 
	Our proposed frameworks can be trained with any SARL and MARL algorithm, and in this study, we selected TD3 and MATD3, which are model-free, off-policy, actor-critic algorithms.
    After completing training in our simplified custom environment designed to prioritize computational efficiency, we evaluated the trained policies in a more realistic \textit{PyBullet} environment for \textit{sim-to-sim} verification. 
    This evaluation allows us to assess how well the policies generalize to a more complex and realistic simulation setting.
    The code of our experiments can be accessed at \url{https://github.com/fdcl-gwu/gym-rotor}.
	
	\textbf{Network Structures.}
	In our experiments, both critic and actor networks were implemented as MLPs with two hidden layers with ReLU activations.
	The critic networks employed 512 nodes per hidden layer for all frameworks.
	In contrast, the actor networks utilized a variable number of hidden nodes depending on the agent: 16 nodes for SARL and Agent$\#$1 in MARL frameworks, while 8 nodes for Agent$\#$2 due to its smaller input dimension.
	Lastly, the actor network's output layer used the hyperbolic tangent activations to ensure control signals remain within the range of $[-1, 1]$.
	
	\textbf{Hyperparameters.}
	The reward coefficients $k_x$ and $k_{b_1}$ in Section \ref{sec:MA_quad} were set to 7.0 and 2.0, respectively, to minimize tracking errors in position and heading.
	To eliminate steady-state errors, $k_{I_x}$ and $k_{I_{b_1}}$ were chosen as 0.04 and 0.06, with $\alpha = 0.01$ in \eqref{eqn:SS}.
	Also, to achieve smooth and stable control, the penalizing terms were set to $k_v = k_\Omega = k_{\Omega_3} = k_{\omega_{12}} = 0.25$ and $k_{b_3} = 3.5$.
	Note that excessively large penalties prioritized stabilization over position tracking and prevented the quadrotor from moving toward its target.
	The selected hyperparameters are summarized in \Cref{tab:hyper}.
	\begin{table}[b]
		\centering
        \caption{Hyperparameters.}
		\label{tab:hyper}
		\begin{tabular}[t]{lc}
			\toprule
			Parameter & Value\\
			\midrule
			Optimizer & AdamW \\
			Learning rate & $1 \times 10^{-4} \rightarrow 1 \times 10^{-6}$ \\
			Discount factor, $\gamma$ & 0.99 \\
			Replay buffer size & $10^6$ \\
			Batch size & 256 \\
			Maximum global norm & 100 \\
			Exploration noise & 0.3 $ \rightarrow$ 0.05 \\
			Target policy noise & 0.2 \\
			Policy noise clip & 0.5 \\
			Target update interval & 3 \\
			Target smoothing coefficient & 0.005 \\
			\bottomrule
		\end{tabular}
	\end{table}

	\textbf{Training Method.} 
    We trained our models for 1.5 million timesteps on a GPU workstation powered by NVIDIA A100-PCIE-40GB, without pre-training or any auxiliary aid techniques such as PID controllers.
	To encourage exploration, we randomly selected all states of each episode from predefined distributions, e.g., randomly placing the quadrotor in $3 \mathrm{m} \times 3 \mathrm{m} \times 3 \mathrm{m}$ space.
	During training and evaluation, both state and action were normalized to $[-1, 1]$ to speed up the training process while improving stability.
	Similarly, we normalized each step's reward to $[0, 1]$ to ensure convergence.
	The learning rate was scheduled with SGDR~\cite{loshchilov2016sgdr} to avoid over-fitting, and the exploration noise was linearly decayed.

    \textbf{Domain Randomization.}
    RL models trained in simulators often suffer from performance degradation when deployed in different simulation environments or the real world. 
    This arises from discrepancies between the training and testing environments, caused by parametric uncertainties or unmodeled dynamics. 
    Domain randomization is one of the promising methods to bridge the resulting \textit{sim-to-sim} or \textit{sim-to-real gap}, where the properties of the training environment are randomized to allow the agent to adapt and generalize across varying conditions \cite{molchanov2019sim}.
	In this paper, at the beginning of each training episode, the physical parameters were randomly sampled from a uniform distribution within a $\pm 5 \%$ range around the nominal values listed in \Cref{tab:params}.
	\begin{table}[t]
		\centering
		\caption{Quadrotor nominal parameters.}
		\label{tab:params}
		\begin{tabular}[t]{lc}
			\toprule
			Parameter & Nominal Value\\
			\midrule
			Mass, $m$ & 2.15 $kg$ \\
			Arm length, $d$ & 0.23 $m$ \\
			Moment of inertia, $J$ & (0.022, 0.022, 0.035) $kg~m^2$ \\
			Torque-to-thrust coefficient, $c_{\tau f}$ & 0.0135 \\
			Thrust-to-weight coefficients, $c_{tw}$ & 2.2 \\
			\bottomrule
		\end{tabular}
	\end{table}

    
	 \subsection{Benchmark Results} 
\begin{figure}[b]
		\centering
		\includegraphics[width=0.485\textwidth]{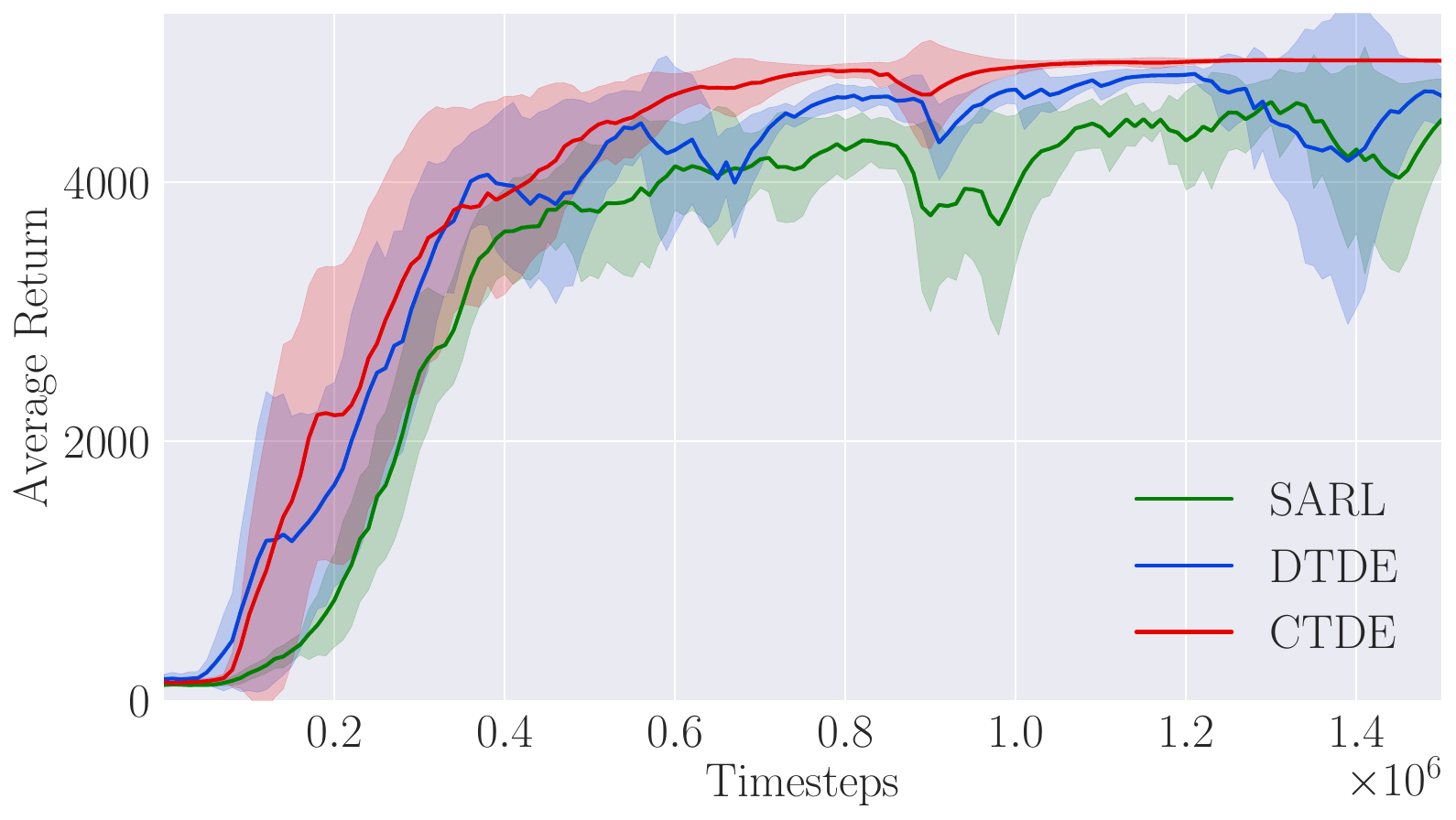}
		\caption{Learning curves for each framework over timesteps.
		The solid lines and shaded areas represent the average and standard deviation of performance, respectively.}
		\label{fig:BM}
	\end{figure}

	 To benchmark the performance of each framework, we evaluated our trained policies at every 10,000 steps without exploration noise.
	 The resulting learning curves, depicting the average return over 5 random seeds, are presented in \Cref{fig:BM}.
	 These curves are smoothed by exponential moving averages with a smoothing weight of 0.8, where the solid lines and shaded areas represent the average and its 2$\sigma$ bounds over random seeds, respectively. 

	 First, our proposed MARL frameworks, namely DTDE and CTDE, exhibit faster and more robust convergence compared to the traditional single-agent framework, SARL. 
	 This highlights the effectiveness of the decoupled dynamics in MARL, as each agent focuses on its unique task, accelerating the learning process and achieving greater stability.

    Next, we compare the performance of DTDE and CTDE.
	In the initial training phase, both frameworks perform similarly, achieving a higher reward relatively faster than SARL.
	However, CTDE shows more consistent training afterward, which becomes more evident as the number of timesteps increases. 
	Notably, in DTDE settings, the second agent for yawing control tends to generate an aggressive torque $M_3$, being too focused on its own goal.
	We hypothesize that the primary cause of this behavior is non-stationarity.
	In single-agent settings, the agent's decision exclusively affects the environment, and state transitions can clearly be attributed to the agent, while external factors are considered part of the system dynamics. 
	However, in multi-agent scenarios, the environment changes dynamically as multiple agents update their policies individually during learning, violating the Markov property \cite{busoniu2008comprehensive}.
	Due to the absence of information exchange between the two agents in DTDE, they struggle to coordinate their actions, leading to suboptimal performance.
	In CTDE, on the other hand, the two agents share their states and actions through the centralized critic networks, enabling better coordination of their decisions.

	\subsection{Simulation Experiments}	
	
	\begin{table}[t]
		\centering
		\renewcommand{\arraystretch}{1.25} 
		\caption{Performance comparison of each RL framework after 1.5M steps of training.
			Average position errors ($\bar{e}_x$ in centimeters) and average heading errors ($\bar{e}_{b_1}$ in degrees) are measured under varying initial yaw errors $\psi_0$.}
		\label{tab:avg_ex_eR}
		\begin{tabular}{c|cccc?cccc}
			\hline
			\multirow{3}{*}{Policy} & \multicolumn{4}{c?}{Simplified Training Env.} & \multicolumn{4}{c}{Sim-to-Sim PyBullet Env.} \\ \cline{2-9} 
			& \multicolumn{2}{c|}{$\psi_0 = 30^{\circ}$} & \multicolumn{2}{c?}{$\psi_0 = 150^{\circ}$} & \multicolumn{2}{c|}{$\psi_0 = 30^{\circ}$} & \multicolumn{2}{c}{$\psi_0 = 150^{\circ}$} \\ \cline{2-9} 
			& \multicolumn{1}{c|}{$\bar{e}_{x}$} & \multicolumn{1}{c|}{$\bar{e}_{b_1}$} & \multicolumn{1}{c|}{$\bar{e}_{x}$} & $\bar{e}_{b_1}$ & \multicolumn{1}{c|}{$\bar{e}_{x}$} & \multicolumn{1}{c|}{$\bar{e}_{b_1}$} & \multicolumn{1}{c|}{$\bar{e}_{x}$} & $\bar{e}_{b_1}$ \\ \hline
			SARL & \multicolumn{1}{c|}{35.0} & \multicolumn{1}{c|}{12.2} & \multicolumn{1}{c|}{40.4} & \multicolumn{1}{c?}{136.7} & \multicolumn{1}{c|}{34.3} & \multicolumn{1}{c|}{12.5} & \multicolumn{2}{c}{crashed} \\
			DTDE & \multicolumn{1}{c|}{12.0} & \multicolumn{1}{c|}{2.10} & \multicolumn{1}{c|}{12.0} & 13.8 & \multicolumn{1}{c|}{12.2} & \multicolumn{1}{c|}{2.19} & \multicolumn{1}{c|}{14.4} & 16.9 \\
			CTDE & \multicolumn{1}{c|}{\textbf{5.81}} & \multicolumn{1}{c|}{\textbf{1.81}} & \multicolumn{1}{c|}{\textbf{5.81}} & \textbf{13.6} & \multicolumn{1}{c|}{\textbf{5.90}} & \multicolumn{1}{c|}{\textbf{1.89}} & \multicolumn{1}{c|}{\textbf{10.1}} & \textbf{16.0} \\ 
			\hline
		\end{tabular}
	\end{table}

	Table \ref{tab:avg_ex_eR} compares the flight performance of each framework through simulation experiments, showcasing their average position $\bar{e}_x$ and heading $\bar{e}_{b_1}$ errors in both training and evaluation environments.
	This highlights their strengths and limitations, particularly regarding their ability to handle various yaw errors.	
	The task involved reaching a target position of $x_d = [0, 0, 0]^T$ from 100 random starting points within a 50 $\mathrm{cm}^3$ cubic space, while simultaneously converging to a desired heading direction of $b_{1_d} = [1,0,0]^T$ with initial yaw errors of 30 and 150 degrees.
	
	We first evaluated each framework in the simplified training environment where the agents are trained.
	In SARL, due to its tightly coupled nature, both position and heading errors increased as yaw errors grew.
	Conversely, both DTDE and CTDE exhibited robust control performance regardless of yaw errors due to their decoupled control strategy.
	Next, we further validated their performance through sim-to-sim transfer in a more realistic physical environment, \textit{gym-pybullet-drones}~\cite{panerati2021learning}.
	For small yaw errors, flight performance in the training and transferred environments remained comparable. 
	However, for larger yaw errors, sim-to-sim experiments showed performance degradation, likely due to the wider domain gap.
	Notably, the quadrotor trained with SARL even crashed at large yaw errors.
	This presents the limitations of domain randomization and highlights the potential for future research on domain adaptation techniques.
		
\begin{figure*}[t] 
		\centering
		\begin{tabular}{cc}
			\subfigure[]{\includegraphics[width=0.46\textwidth]{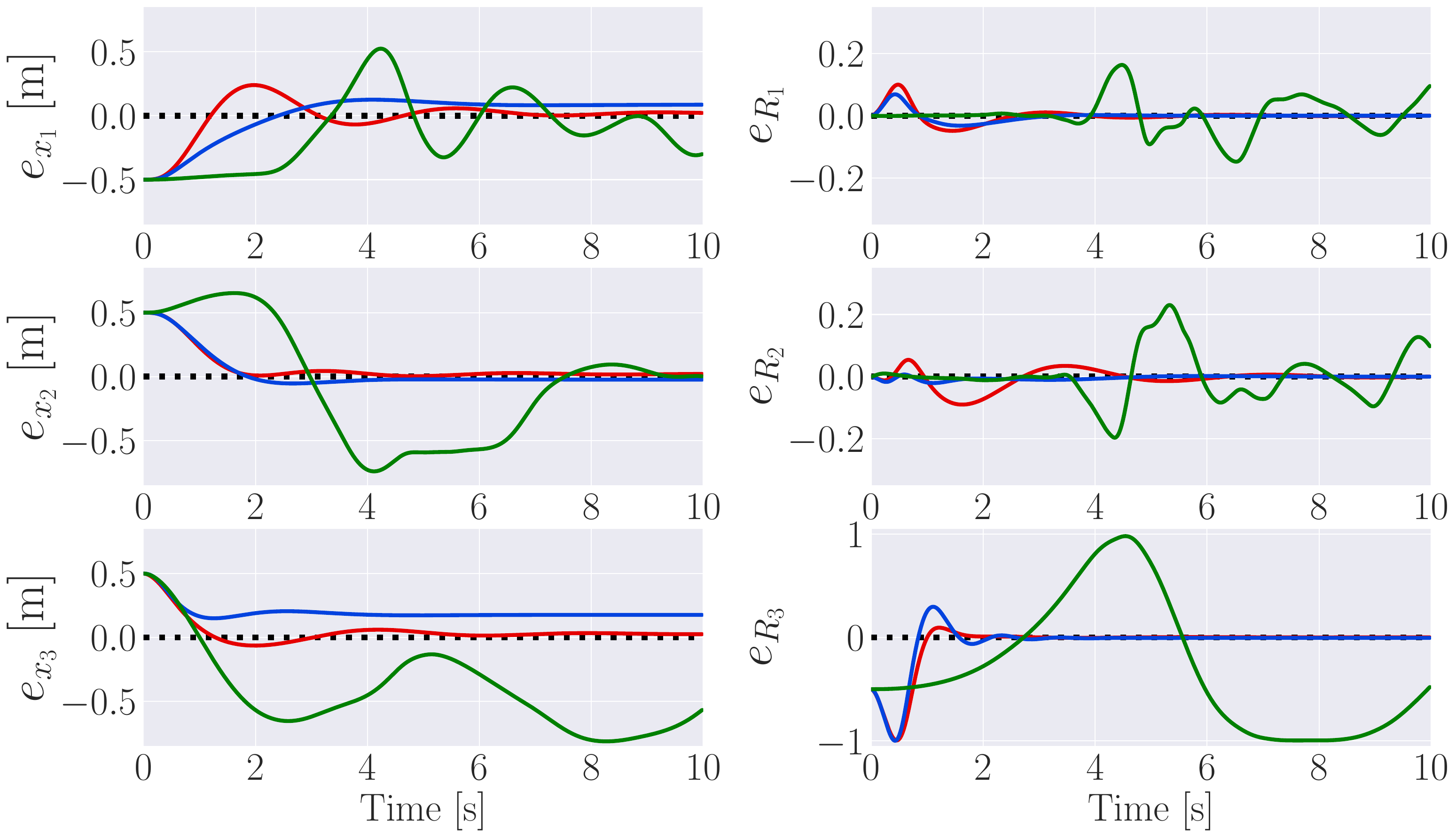} \label{fig:ex_eR_sim}} & \hspace{0.1em}
			\subfigure[]{\includegraphics[width=0.46\textwidth]{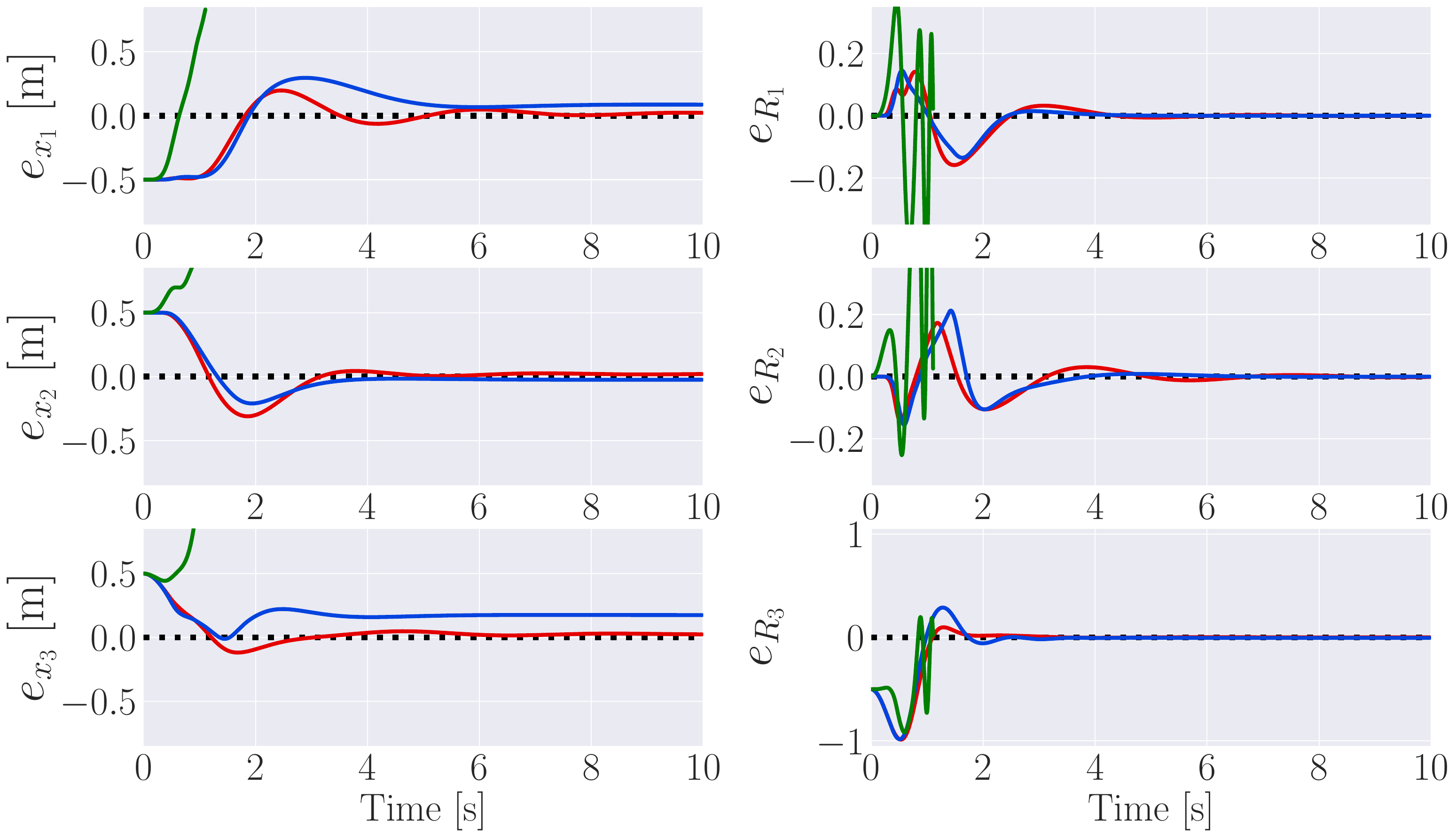} \label{fig:ex_eR_pybullet}} \\[5pt]
		\end{tabular}
		\caption{Flight performance comparison of SARL (green), DTDE (blue), and CTDE (red) in a single episode when $e_{b_1}(0) = 150^{\circ}$ under the same initial conditions.
			(a) simulation in the training environment (b) simulation in the \textit{PyBullet}-based physics environment, which represents the sim-to-sim transfer.
			Each column visualizes (left) position error, $e_x = x - x_d \in \Re^3$, (right) attitude error, $e_R = \frac{1}{2} (R_d^T R - R^T R_d)^\vee \in \Re^3$.}
		\label{fig:ex_eR_traj}
	\end{figure*}

	Finally, \Cref{fig:ex_eR_traj} presents the flight trajectories controlled by each framework, particularly focusing on their ability to recover from a large initial yaw error of 150 degrees.
	The position and attitude errors of our proposed methods, DTDE and CTDE, are denoted by blue and red lines, respectively.
	For comparison, the trajectories of SARL are represented by green lines.
	While SARL failed to stabilize at the goal position and exhibited large overshoots and oscillations, both DTDE and CTDE agents achieved satisfactory flight performance by separating yaw control from roll and pitch control, highlighting the benefits of the decoupled approach.
	However, compared to CTDE, DTDE exhibited larger position steady-state errors, requiring at least 2M training steps to eliminate these errors.
	Therefore, our results consistently demonstrate a clear advantage of centralized training over decentralized training in both training efficiency and flight performance.
	
	\section{Conclusion} 	
	In this paper, we present multi-agent reinforcement learning frameworks for quadrotor low-level control by decomposing the quadrotor dynamics into translational and yawing parts, each exhibiting distinct characteristics.
	Additionally, we introduce regularization terms to mitigate steady-state errors and prevent aggressive control inputs, promoting efficient control.
	We demonstrate that our MARL frameworks outperform traditional single-agent counterparts through sim-to-sim verification, significantly enhancing training efficiency while maintaining flight-worthy control performance. 

	While simulation verification provides a valuable proving ground, our ongoing efforts aim to advance its applicability in real-world scenarios.
	Future work involves validating our frameworks through actual flight experiments, known as \textit{sim-to-real transfer}, using domain randomization or adaptation techniques to bridge the reality gap in multi-agent settings.

	\bibliography{BibTeX}
	\bibliographystyle{IEEEtran}

\end{document}